\documentclass[sigconf]{acmart}
\usepackage{tikz}
\usepackage{multirow}
\usetikzlibrary{shapes.geometric, arrows}
\AtBeginDocument{%
  \providecommand\BibTeX{{%
    \normalfont B\kern-0.5em{\scshape i\kern-0.25em b}\kern-0.8em\TeX}}}

\setcopyright{acmcopyright}
\copyrightyear{2018}
\acmYear{2018}
\acmDOI{XXXXXXX.XXXXXXX}

\acmConference[The Web Conference 2024]{Singapore}
\acmPrice{Singapore}
\acmISBN{978-1-4503-XXXX-X/18/06}

\begin{document}

\title{Enhanced E-Commerce Attribute Extraction: Innovating with Decorative Relation Correction and LLAMA 2.0-Based Annotation}

\author{Jianghong Zhou}
\authornotemark[1]
\email{{jianghong.zhou, weizhi.du,- mdomarfaruk.rokon}@walmart.com}
\orcid{0009-0006-1720-1425}
\author{Weizhi Du}
\orcid{0000-0001-7448-8190}
\author{Md Omar Faruk Rokon}
\orcid{0000-0002-1385-9389}
\affiliation{%
  \institution{Walmart Global Tech}
    \streetaddress{ 860 W California Ave.}
  \city{Sunnyvale}
  \state{California}
  \country{USA}
  \postcode{94086}
}

\author{Zhaodong Wang}
\author{Jiaxuan Xu}
\author{Isha Shah}
\orcid{0009-0001-6498-7954}
\email{{Zhaodong.Wang, Jiaxuan.Xu,- Isha.Shah@walmart.com}@walmart.com}
\affiliation{%
  \institution{Walmart Global Tech}
    \streetaddress{ 860 W California Ave.}
  \city{Sunnyvale}
  \state{California}
  \country{USA}
  \postcode{94086}
}

\author{Kuang-chih Lee}
\author{Musen Wen}
\email{{Kuang-chih.Lee, musen.wen-@walmart.com}@walmart.com}
\affiliation{%
  \institution{Walmart Global Tech}
  \streetaddress{ 860 W California Ave.}
  \city{Sunnyvale}
  \state{California}
  \country{USA}
  \postcode{94086}
}

\renewcommand{\shortauthors}{Jianghong Zhou, et al.}

\begin{abstract}
The rapid proliferation of e-commerce platforms accentuates the need for advanced search and retrieval systems to foster a superior user experience. Central to this endeavor is the precise extraction of product attributes from customer queries, enabling refined search, comparison, and other crucial e-commerce functionalities. Unlike traditional Named Entity Recognition (NER) tasks, e-commerce queries present a unique challenge owing to the intrinsic decorative relationship between product types and attributes. In this study, we propose a pioneering framework that integrates BERT for classification, a Conditional Random Fields (CRFs) layer for attribute value extraction, and Large Language Models (LLMs) for data annotation, significantly advancing attribute recognition from customer inquiries. Our approach capitalizes on the robust representation learning of BERT, synergized with the sequence decoding prowess of CRFs, to adeptly identify and extract attribute values. We introduce a novel decorative relation correction mechanism to further refine the extraction process based on the nuanced relationships between product types and attributes inherent in e-commerce data. Employing LLMs, we annotate additional data to expand the model's grasp and coverage of diverse attributes. Our methodology is rigorously validated on various datasets, including Walmart, BestBuy's e-commerce NER dataset, and the CoNLL dataset, demonstrating substantial improvements in attribute recognition performance. Particularly, the model showcased promising results during a two-month deployment in Walmart's Sponsor Product Search, underscoring its practical utility and effectiveness. This initiative not only lays a robust foundation for augmenting e-commerce search and retrieval systems but also delineates a scalable approach to accommodate a broad array of attributes by leveraging LLMs for data annotation, adeptly catering to the evolving demands of e-commerce landscapes.
\end{abstract}




\maketitle

\section{Introduction}

The rapid proliferation of e-commerce platforms has catalyzed a transformative shift in retail, offering consumers an unprecedented array of choices and the convenience of shopping from anywhere, at any time. This digital marketplace, burgeoning with extensive product listings and intricate user queries, demands an equally sophisticated search experience. Central to such an experience is the ability to precisely recognize product attributes from customer inquiries—a cornerstone for not only accurate search results but also for the nuanced comparison and facilitation of essential e-commerce operations  \cite{yadav2019survey}. High-precision attribute recognition is not just a technical aspiration; it is the linchpin for mirroring user intent, thereby elevating user satisfaction and enhancing the overall usability of e-commerce platforms.

\begin{figure}[h]
  \centering
  \includegraphics[width=\linewidth]{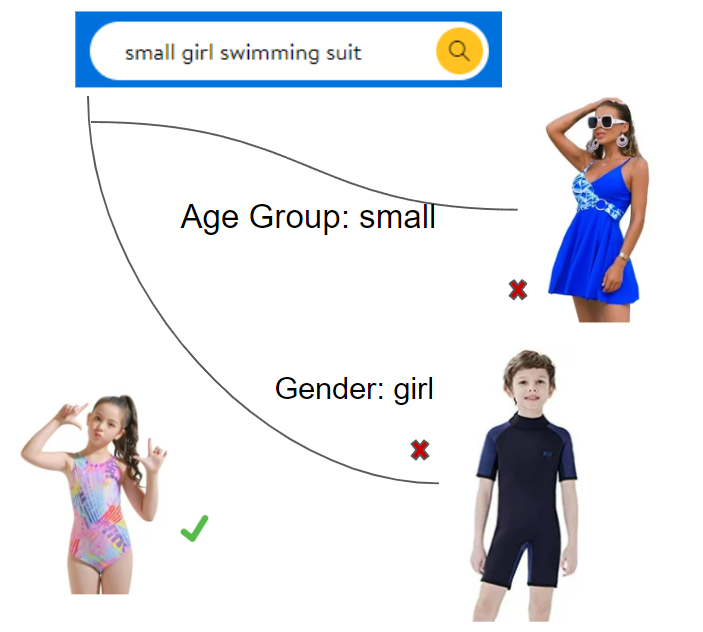}
  \caption{A case study of query: small girl swimming suit}
  \label{case}
\end{figure}

To elucidate, consider a case study illustrated in Fig \ref{case}, where a user searches for a "small girl swimming suit". Effective attribute recognition can discern that the user is seeking a swimsuit for a small girl, thereby filtering out irrelevant products such as "Sexy Two Piece Swimdress Swimsuits for Women Girl with Short Skirt Bottom V-Neck" and "Kids Wetsuit for Boys", which may otherwise appear in the search results.

Upon examining our case study, we encounter several technical challenges that current attribute recognition systems must surmount. Firstly, the scalability of these systems is tested as e-commerce inventories burgeon, demanding a flexible and adaptive approach to handle the growing variety of user queries. Secondly, the systems must navigate the intricacies of natural language to avoid contextual ambiguities that lead to misclassification. Thirdly, the specialized jargon inherent to diverse e-commerce products often eludes the grasp of generic NER systems, necessitating a more nuanced approach to domain-specific language. Fourthly, the computational heft of LLMs poses a hurdle for their practical application in real-time tasks. Finally, the reliability of human-annotated datasets for training these systems is frequently questionable, affecting the overall robustness of attribute recognition \cite{wen2019building}. These challenges collectively underscore the need for an innovative solution capable of real-time processing and high accuracy to significantly enhance the e-commerce search experience \cite{TolokaAI, he2023annollm,carriereimproving}.

Building on the foundation laid out by the identified challenges, our proposed solution employs a multifaceted machine-learning approach to revolutionize attribute recognition within e-commerce platforms. At the heart of our solution is the BERT model, lauded for its ability to effectively capture the contextual relationships between words, thereby proving invaluable in text classification tasks \cite{qiu2022pre}. Complementing BERT, we integrate Conditional Random Fields (CRFs), renowned for their sequence labeling capabilities that are particularly well-suited for Named Entity Recognition (NER)—a task that shares a close affinity with attribute recognition \cite{liu2020chinese}.

To enhance our methodology, we acknowledge the constraints posed by the latency and computational demands of Large Language Models (LLMs), which can be challenging for real-time industry applications such as search, retrieval, and bidding systems that necessitate swift intent detection \cite{zhou2023leveraging}. Despite these challenges, we strategically employ LLMs for data annotation, tapping into their vast capabilities to broaden the spectrum of detectable attributes. This judicious application of LLMs significantly augments our model's comprehension and its ability to adapt to a wide variety of product queries, thereby improving performance in e-commerce environments where an accurate and quick understanding of user intent is paramount \cite{wang2023survey}.

A cornerstone of our framework is the novel decorative relation correction mechanism. This mechanism is adept at deciphering the often intricate relationship between product types and attributes, a challenge peculiar to the e-commerce context \cite{kim2002product, zhou2020diversifying}. By incorporating this mechanism, our model can accurately interpret queries such as "red leather iPhone 12 case," ensuring that it understands the user's intent to find a case with specific attributes—red in color, leather in texture, for an iPhone 12—rather than an iPhone itself of that description. Similarly, for a search like "stainless steel 30-inch refrigerator," our model applies the decorative relation to focus the search on refrigerators that match the given attributes of material and size, effectively sifting through the vast inventory to present the user with the most relevant products. These examples highlight the transformative potential of the decorative relation in enhancing the precision and efficiency of the attribute recognition process.

We empirically validate our approach on several datasets, including Walmart \cite{akande2021application}, BestBuy's e-commerce NER dataset \cite{sidorov2018attribute}, and the CoNLL dataset \cite{carreras2005introduction}. The results exhibit notable improvements in attribute recognition performance, particularly demonstrating promising results during a two-month deployment in Walmart's Advertising Retrieval. Furthermore, the scalability of our model is showcased through the ease of extending recognizable attributes by employing LLMs for data annotation.

Our contributions are fourfold:
\begin{itemize}
\item We unveil a novel framework amalgamating BERT, CRFs, and LLMs to fortify attribute recognition, crucial for reflecting users' intent in e-commerce searches and related tasks.
\item We introduce an innovative decorative relation correction mechanism to harness the inherent relationship between product types and attributes in e-commerce, further refining the attribute extraction process.
\item We empirically validate our approach on diverse datasets, including Walmart, and BestBuy's e-commerce NER dataset, and the CoNLL dataset, illustrating significant improvements in attribute recognition performance.
\item Our model displayed encouraging results during a two-month deployment in Walmart's Sponsor Product Search, underscoring its practical utility and effectiveness.
\item We elucidate the scalability of our model by leveraging LLMs for data annotation, offering a pathway to accommodate an expansive array of attributes with diminished human intervention.
\end{itemize}

The rest of the paper is organized as follows: Section 2 reviews related work in attribute recognition, NER, and LLMS-based data annotation. Section 3 elucidates the proposed framework. Section 4 delineates the experimental setup, results, and discussion. Section 5 concludes the paper and demarcates future directions.

\section{Related Work}

\subsection{Named Entity Recognition (NER)}
Named Entity Recognition (NER) remains a pivotal task in information extraction, where the objective is to categorize named entities into predefined groups such as persons, organizations, and locations. Recent advancements in NER have explored various objective functions like cross entropy and Conditional Random Fields (CRFs) to optimize model performance, particularly in scenarios with balanced data distributions and sufficient annotated training examples \cite{nguyen2023auc,zhou2020rlirank}.

\subsection{Conditional Random Fields (CRFs)}
CRFs, a probabilistic sequence labeling model, has been extensively utilized in NER tasks. It has shown efficacy in identifying and classifying named entities, especially in morphologically rich languages where statistical algorithms achieve good identification and classification accuracy, albeit requiring additional knowledge to further improve accuracy. For instance, Stanford CoreNLP's CRFs model has demonstrated substantial improvements in NER by accurately identifying and categorizing entities in text \cite{finkel2005incorporating, zhou2020diversifying}.

\subsection{Llama 2.0}

Llama 2.0 marks an evolutionary leap within the landscape of large language models (LLMs), presenting an advanced collection of models that have been meticulously pre-trained and subsequently fine-tuned to excel in dialogue-oriented tasks. This progression has been thoughtfully documented by \cite{touvron2023llama}, who underline the significant enhancements that Llama 2.0 brings to the table. The development of Llama 2.0 reflects a growing trend in the artificial intelligence sector towards embracing open-source frameworks. Such a movement is instrumental in democratizing AI advancements, paving the way for widespread academic and commercial applications.

Notably, Meta positions Llama 2.0 at the forefront of this transformative wave, characterizing it as a trailblazer among next-generation open-source LLMs. The implications of this for e-commerce are particularly profound. In e-commerce settings, where understanding and generating human-like dialogue is paramount, Llama 2.0 can provide unprecedented capabilities. Whether it's powering sophisticated chatbots that can handle complex customer service interactions or enhancing search algorithms to understand and respond to nuanced user queries, Llama 2.0 offers a new level of interactional finesse \cite{zhou2023leveraging, zhou2017block, zhou2020diversifying}.

\subsection{BERT}

The introduction of BERT (Bidirectional Encoder Representations from Transformers) has revolutionized the domain of Named Entity Recognition (NER). This groundbreaking model, introduced by \cite{devlin2018bert}, heralded a new era in NLP by leveraging bidirectional training on an extensive corpus to enrich the understanding of textual context. Such contextual comprehension is particularly critical for NER, where the meaning and classification of words are deeply entwined with their surrounding text \cite{sun2021rpbert}.

BERT has markedly surpassed previous models in NER tasks, which traditionally leaned on unidirectional textual analysis and extensive feature engineering. Its pre-training on voluminous and varied text provides deep, context-rich word representations that have consistently achieved superior precision and recall in entity identification, frequently establishing new performance standards \cite{pires2019multilingual}.

The ingenuity of BERT prompted the development of variants like RoBERTa \cite{liu2019roberta}, DistilBERT \cite{sanh2019distilbert}, and ALBERT \cite{lan2019albert}, each designed to optimize aspects like computational efficiency and model size without sacrificing performance. These adaptations are crucial for practical applications where swift and efficient processing is paramount.

In the e-commerce sector, BERT and its derivatives have excelled in interpreting product details and customer feedback, enabling the extraction of entities that are pertinent to search queries and recommendation engines. The model's nuanced understanding of language nuances significantly enhances the accuracy and relevance of search results, aligning them more closely with user expectations \cite{xie2020sentiment,zhou2021biased}.

In summary, BERT's architecture has laid a foundational stone in the NER landscape, steering the direction of future NLP advancements and serving as a benchmark in the field's continuous progression.

\begin{figure*}
  \centering
  \includegraphics[width=\linewidth]{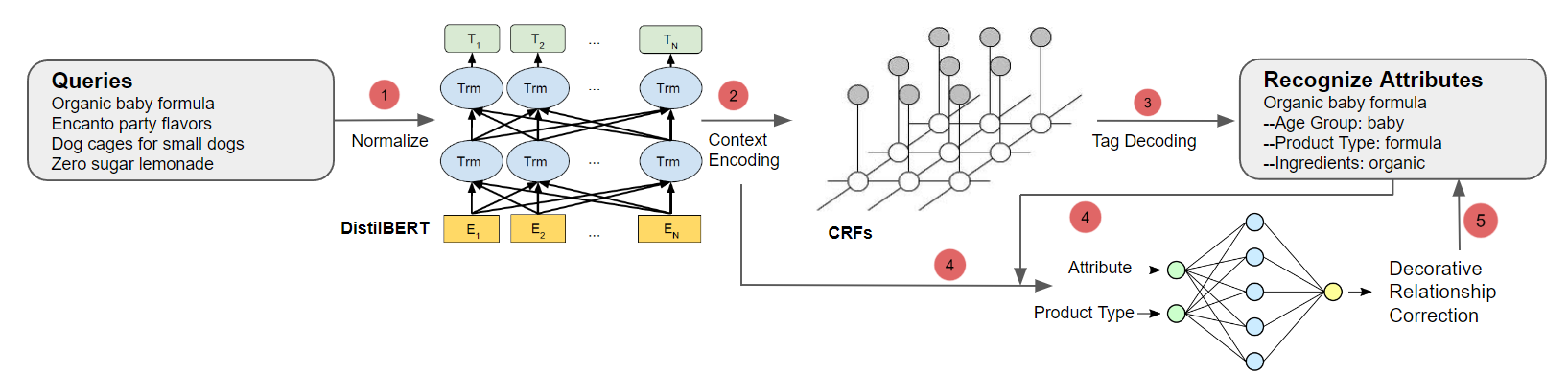}
  \caption{Illustration of the Bert-CRFs NER Method with decorative relationship correction.}
  \label{decor}
\end{figure*}

\section{Methodology}
Our methodology encapsulates three pivotal components: BERT-CRFs NER, Decorative Relation Correction, and Attribute Expansion using LLAMA 2.0. Each of these components is meticulously designed to address specific challenges in attribute recognition from customer queries in e-commerce scenarios. In the ensuing subsections, we delineate each component in detail. The whole process of the attribute extraction process is summarized in \ref{decor}.

\subsection{BERT-CRFs Named Entity Recognition}

In this section, we delve into the details of encoding and tag decoding within our proposed BERT-CRFs framework for Named Entity Recognition (NER).

\subsubsection{Encoding}
The encoding phase is crucial for capturing contextual information from the input text, which in turn aids in accurate attribute recognition. We employ BERT to generate contextual embeddings for each token in the input text. BERT processes the input text and outputs a sequence of contextual embeddings, where each embedding is a high-dimensional vector representing a token in the context of its surrounding tokens.

The encoding process can be formally represented as follows:
\begin{equation}
    E = \text{BERT}(x)
\end{equation}
where \( x \) is the input text and \( E \) is the sequence of contextual embeddings.

\subsubsection{Tag Decoding}
Following the encoding phase, the generated contextual embeddings are fed into a CRFs layer for tag decoding. The CRFs layer operates on the sequence of embeddings to produce a sequence of tags, each corresponding to a token in the input text, indicating whether the token is part of an attribute and the type of attribute it belongs to.

The objective of the CRFs layer is to find the most probable sequence of tags \( y \) given the sequence of contextual embeddings \( E \). This can be modeled as:

\begin{equation}
    y^* = \arg\max_y P(y|E)
\end{equation}

The probability \( P(y|E) \) is computed using the following formula, derived from the definition of a linear-chain CRF:

\begin{equation}
    P(y|E) = \frac{1}{Z(E)} \exp\left(\sum_{i=1}^n \theta_f f_i(E, y)\right)
\end{equation}
where \( Z(E) \) is the normalization factor, \( \theta_f \) are the learned feature weights, and \( f_i \) are feature functions that capture the relationship between the input embeddings and the output tags, as well as the transition relationships between adjacent tags.

The feature functions \( f_i \) and the feature weights \( \theta_f \) are learned from the training data to minimize the negative log-likelihood of the true tag sequences.

This two-step process of encoding and tag decoding facilitates the extraction of attribute values from customer queries by first capturing the contextual information using BERT and then employing a CRFs layer to decode the most probable sequence of tags that indicate the attributes present in the queries.

\subsection{Decorative Relation Correction}
Utilizing the encodings obtained from BERT, we construct a binary classification neural network to ascertain the decorative relationship between product types and attributes. Given a pair consisting of a product type and attributes extracted from the NER model, we employ their encodings as input to the neural network. The network is trained to discern whether a decorative relationship exists between the pair. The model can be mathematically represented as:

\begin{equation}
    y = \sigma(W \cdot [e_{\text{attr}}, e_{\text{ptype}}] + b)
\end{equation}

where \( e_{\text{attr}} \) and \( e_{\text{ptype}} \) denote the encodings of the attribute and product type respectively, \( W \) and \( b \) represent the weight matrix and bias term, and \( \sigma \) denotes the sigmoid activation function. Attributes that do not adhere to a decorative relationship with their respective product types are identified by suboptimal results from the NER model. The neural network comprises three dense layers.

To illustrate, consider the case of "tahini sauce for hummus". Here, two situations arise: considering 'sauce' as the product type or considering 'hummus' as the product type. For instance, two recognized results could be (Flavor: tahini, 'Cuisine Type': sauce, Product Type: Hummus) and a suboptimal result with a lower score (Flavor: tahini, 'Product Type: sauce', "Flavor: Hummus"). Initially, we check the pair of attributes and product type, examining (tahini, Hummus), which is valid, but (sauce, hummus) is not decorative, yielding a score of 1. Subsequently, we check the next answer, where both (tahini, sauce) and (Hummus, sauce) are valid, resulting in a score of 2. Hence, we select the second answer based on a higher score.

The protocol entails utilizing the decorative relation correction to scrutinize the top 3 answers, and return the highest scoring answer if a product type is recognized (noting that the product type is also one of the attributes).

\subsection{Attribute Expansion using LLAMA 2.0}
\begin{figure}[h]
  \centering
  \includegraphics[width=\linewidth]{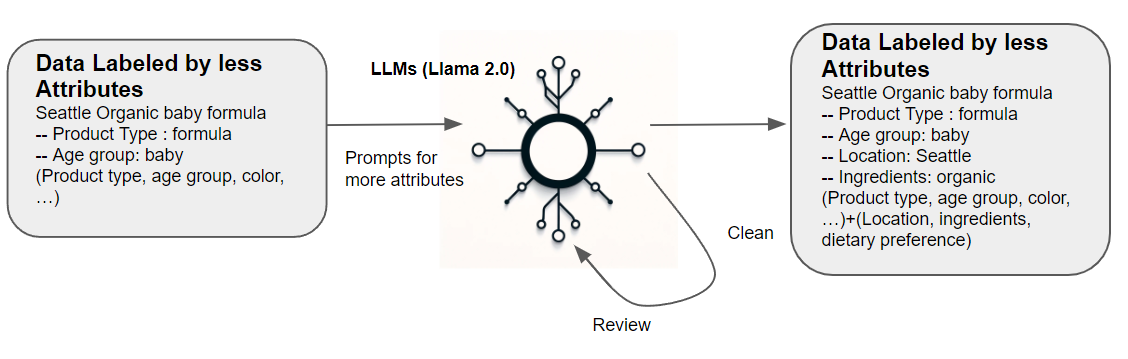}
  \caption{Process of LLMs Attribute Expansion}
  \label{llm}
\end{figure}

To augment the range of recognizable attributes, we employ LLAMA 2.0 to generate prompts soliciting additional attributes. The prompts are meticulously crafted and fed into LLAMA 2.0, which in turn generates responses containing potential attributes. A cleaning process follows to eliminate hallucinations and incorrect format responses. Subsequently, LLAMA 2.0 is utilized once more to review the results, enhancing the accuracy of the annotations. This iterative process of attribute solicitation, cleaning, and reviewing significantly extends the spectrum of recognizable attributes, making our approach robust and scalable. The process is illustrated in Fig \ref{llm}.

\begin{equation}
    A = \text{LLAMA}_{\text{ask}}(\text{Prompt}(q))
\end{equation}

\begin{equation}
    A_{\text{new}} = \text{LLAMA}_{\text{review}}(\text{Clean}(A))
\end{equation}

In the process delineated, $q$ is the customer query and \( A \) signifies the ensemble of attributes initially discerned by the model. The function \( \text{Prompt} \) is tasked with the generation of prompts that are instrumental in the attribute recognition workflow. The term \( \text{LLAMA}_{\text{ask}} \) designates the attribute generation phase of the LLAMA 2.0 system, which processes the prompts to procure a preliminary attribute set. Subsequently, \( A_{\text{new}} \) is derived as the result of the \( \text{LLAMA}_{\text{review}} \) phase, wherein the \( \text{Clean} \) function is employed. This crucial function serves to refine the initial attribute set by eliminating any erroneous or fictitious attributes that do not have a basis in the actual dataset.

\section{Experiments}
The objectives of our experiment are threefold: 
\begin{enumerate}
    \item To validate the performance of our proposed method on the e-commerce dataset.
    \item To compare our model's performance against commonly used baseline models.
    \item To investigate the model's robustness and effectiveness when expanded to accommodate a broader set of attributes.
\end{enumerate}

\subsection{Dataset}
We evaluate our proposed framework on several benchmark datasets including the Walmart, BestBuy's e-commerce NER dataset, and the CoNLL dataset. The datasets encompass a diverse range of product categories and attributes, providing a comprehensive ground for evaluating the performance and robustness of our approach.

\begin{itemize}
    \item \textbf{Walmart Dataset \cite{akande2021application}:} This dataset comprises product listings from Walmart's online platform, annotated with product types and attributes across various categories such as electronics, clothing, and home appliances. It contains around 140,000 queries identified as hard negatives, which are likely to yield comparatively low Click Through Rates (CTR) in searches. Initially, attributes and product types were obtained from Walmart's proprietary APIs, annotated by human annotators, sellers, or classifier models. The dataset originally included attributes such as "Brand," "Color," "Size," "Gender," "Age Group," "Deal," and "Product Type." We employed LLAMA 2.0 to expand the attribute set to include "Material," "Price Range," "Condition," "Location," "Ingredients," "Dietary Preferences," "Cuisine Type," "Flavor," "Nutritional Information," "Package" alongside the existing attributes.
    
    \item \textbf{BestBuy's e-commerce NER Dataset \cite{sidorov2018attribute}:} This dataset contains product listings from BestBuy, annotated with product types and attributes, primarily covering electronics, appliances, and other consumer goods. Both the Walmart and BestBuy datasets were merged to form an extended dataset post deduplication, resulting in about 4,000 records with 50\% of these records expertly annotated. For instance, in the short description "Apple watch series 3 42mm from \$339," experts annotated "Apple" as "Brand" and "watch" as "Category". "Category" is considered equivalent to "Product Type" in our experiment.
    
    \item \textbf{CoNLL Dataset \cite{carreras2005introduction}:} Although not tailored for e-commerce, the CoNLL dataset serves as a well-established benchmark for evaluating the performance of our NER models in a generalized setting. The CoNLL-2003 shared task focuses on language-independent named entity recognition, targeting four types of named entities: persons, locations, organizations, and miscellaneous entities. The dataset contains around 21,000 data points, formatted in four columns separated by a single space, with each word on a separate line and an empty line after each sentence. The first item on each line is a word, followed by a part-of-speech (POS) tag, a syntactic chunk tag, and the named entity tag. The dataset employs the IOB2 tagging scheme, distinguishing it from the original dataset which uses IOB1.Our decorative mechanism operates on the dataset by evaluating each entity identified as a product type to calculate the decorative relationship. We compute scores for these relationships and select the one with the highest score from the top three results as the definitive outcome.
\end{itemize}

\subsection{Implementation Details}
Our framework is implemented using the PyTorch library. For the BERT-CRFs model, we utilize the BERT base model pre-trained on the BookCorpus and English Wikipedia datasets. The CRFs layer is implemented using the CRFs suite available in the AllenNLP library. The training is carried out on a machine equipped with an NVIDIA Tesla P100 GPU.

\begin{itemize}
    \item \textbf{Model Training:} The model is trained using the Adam optimizer with a learning rate of \(1e-5\). We use a batch size of 32 and train the model for 5 epochs.
    
    \item \textbf{LLAMA Annotation:} For LLAMA-based data annotation, we employ LLAMA 2.0 to generate prompts and review annotation results. The prompts are meticulously designed to extract precise attribute values from the LLMs. We leverage the LLAMA 2.0 13B chat hf version model for this purpose.

    \item \textbf{Decorative Relation Correction:} The neural network utilized for the decorative relation correction is a straightforward feed-forward network with a single hidden layer. This network is trained separately on pairs of tokens derived from the Walmart dataset. A pair is labeled as positive if it consists of an (attribute, product type) combination; otherwise, it is labeled as negative.

\end{itemize}

\subsection{Prompt Engineering}
In order to acquire more attributes and validate them, we engineered prompts to elicit accurate responses from the LLMs. The prompts are structured into three parts: assumption, example, and question. Below, we detail the prompts used for attribute expansion and review.

\subsubsection{Attributes Expansion} This prompt is for attributes annotation.

\noindent\textbf{Assumption:} 
You are an excellent e-commerce attribute extractor. You will be asked to identify the attributes appearing in a query from a customer. Give the most accurate answer you can come up with. Give short answers, not long answers. Those attributes should strictly and only be selected from "Brand", "Color", "Size", "Material", "Price Range", "Gender", "Age Group", "Condition", "Location", "Ingredients", "Dietary Preferences", "Cuisine Type", "Flavor", "Nutritional Information", "Deal", "Product Type", "Package".

\noindent\textbf{Example:}
\begin{itemize}
    \item Human: Please extract the attributes from the query 'a large red woman t-shirt'
    \item Your answer: ["Size: large", "Color: red", "Gender: woman", "Product Type: t-shirt"]
    \item Human: Please extract the attributes from the query 'zatrains frozen meal blackened chicken alfredo'
    \item Your answer: ["Brand:zatrains", "Flavor: blackened chicken alfredo", "Product Type: frozen meal"]
\end{itemize}

\noindent\textbf{Question:}
Please extract the existing attributes from the query \{query\} and only answer with formats exactly like ["attribute: value"].

\subsubsection{Review} This prompt is for the annotation answer review.

\noindent\textbf{Assumption:}
You are an excellent e-commerce expert. You will be asked to examine the attributes given in a query from a customer. Those attributes should strictly and only select from "Brand", "Color", "Material", "Price Range", "Gender", "Age Group", "Condition", "Location", "Ingredients", "Dietary Preferences", "Cuisine Type", "Flavor", "Nutritional Information", "Size", "Deal", "Product Type", "Package".

\noindent\textbf{Example:}
\begin{itemize}
    \item Human: Please examine the attributes:{"Size": "large", "Color": "red", "Gender": "woman", "Product Type": "t-shirt"} for the query 'a large red woman t-shirt'
    \item Your answer: True
    \item Human: Please examine the attributes:{"Flavor": "spicy", "Material": "ground", "Ingredients": "beef"} for the query 'spicy ground beef'
    \item Your answer: False
\end{itemize}

\noindent\textbf{Question:}
Please examine the attributes \{answer\} for the query \{query\}.

\subsection{Evaluation Metrics}
In the evaluation phase, the performance of the Named Entity Recognition (NER) system is gauged using metrics that are derived from the count of True Positives (TP), False Positives (FP), and False Negatives (FN). These metrics are defined as follows:

\begin{itemize}
    \item \textbf{True Positive (TP):} An entity that is correctly identified by the NER system and also appears in the ground truth.
    \item \textbf{False Positive (FP):} An entity that is identified by the NER system but does not appear in the ground truth.
    \item \textbf{False Negative (FN):} An entity that is not identified by the NER system but appears in the ground truth.
\end{itemize}

The derived metrics include Precision, Recall, and F-score, which are calculated as follows:

   $ \text{Precision}  = \frac{TP}{TP + FP},$ 
      $ \text{Recall}  = \frac{TP}{TP + FN}, $
   
   $ \text{F-score}  = 2 \times \frac{\text{Precision} \times \text{Recall}}{\text{Precision} + \text{Recall}}.$

\subsection{Baselines}
We explore various combinations of text encoders and tag decoders to construct our baseline models. These methods are mature approaches, with some combinations representing the state-of-the-art (SOTA) in Named Entity Recognition (NER) \cite{li2020survey}. The configurations are divided into three groups based on the tagging and correction mechanisms employed:

\noindent\textbf{Softmax-based Methods:}
\begin{itemize}
\item LSTM-Softmax \cite{jia2020multi}: Utilizing an LSTM network coupled with a softmax activation, this method is designed for cross-domain NER, leveraging the intrinsic correlations of entity types to enhance transfer learning capabilities.
\item GPT-2-Softmax \cite{zheng2021adapting}: By fine-tuning the GPT-2 model with a softmax layer, this method adapts the transformer-based architecture for NER, improving upon traditional models by utilizing the transformer's capability to capture bidirectional context.
\item BERT-Softmax \cite{jia2020entity}: Integrating BERT with a softmax classification layer, this approach excels in entity recognition by pre-training on extensive corpora and then fine-tuning for specific NER tasks, thus capturing deep contextual nuances.
\end{itemize}

\noindent\textbf{CRFs-based Methods:}
\begin{itemize}
\item LSTM-CRFs \cite{ghaddar2018robust}: This method combines LSTM networks with Conditional Random Fields (CRFs) to effectively model the sequence and dependencies between tags in NER, resulting in significant improvements in both precision and recall.
\item GPT-2-CRFs \cite{li2020few}: Extending the GPT-2 architecture with a CRF layer, this approach allows for more refined sequence modeling, particularly advantageous in complex NER scenarios where dependencies between entity labels are strong.
\item BERT-CRFs \cite{dai2019named}: This model brings together the strengths of BERT in understanding context with the sequence modeling capabilities of CRFs, providing a powerful tool for accurate entity recognition across various datasets.
\end{itemize}

\noindent\textbf{DRC-enhanced CRFs Methods:}
\begin{itemize}
    \item \textbf{LSTM-CRFs-DRC:} An enhanced LSTM-CRF model with a Decorative Relations Correction mechanism, improving the extraction of attributes by understanding their interrelations within product descriptions.
    \item \textbf{GPT-2-CRFs-DRC:} A GPT-2 based model augmented with CRFs for sequence prediction and DRC for fine-tuning the recognition of complex entity relationships, elevating attribute discernment.
    \item \textbf{BERT-CRFs-DRC (Proposed Method):} This approach integrates BERT's contextual awareness with CRFs and a DRC component to refine the attribute extraction process, aligning it closely with the nuanced intent of e-commerce search queries.
\end{itemize}

The citation list associated with the baselines includes previous work conducted on other datasets or tasks, showcasing their usage and practicality. The Softmax-based methods employ a Softmax layer for tagging, the CRFs-based methods utilize a Conditional Random Field (CRF) layer for tagging, and the DRC-enhanced CRF methods incorporate a Decorative Relation Correction (DRC) mechanism in addition to a CRFs layer to enhance tagging accuracy.

\begin{figure}
  \centering
  \includegraphics[width=\linewidth]{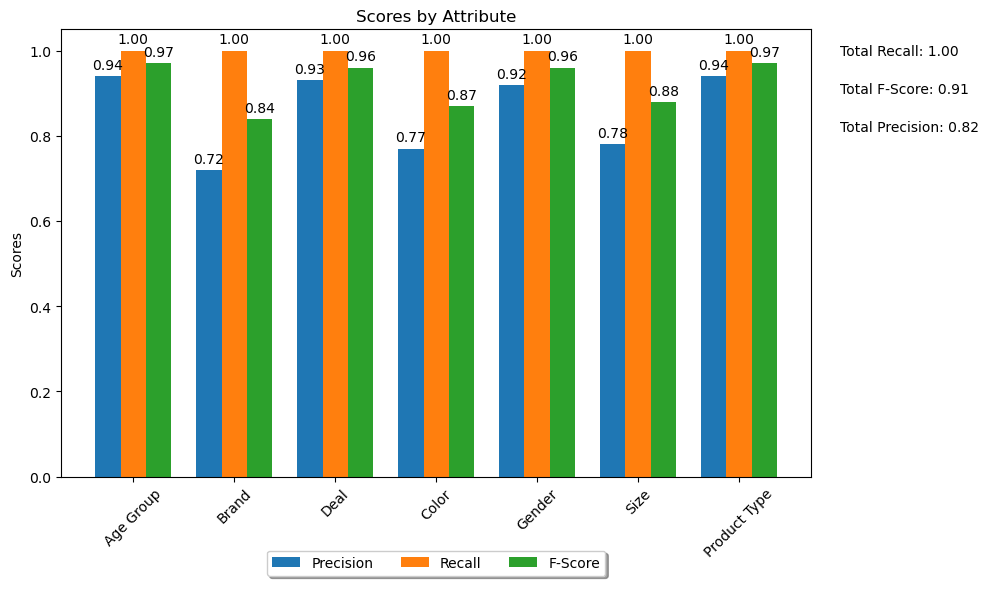}
  \caption{The performance of Bert-CRFs-DRC on the Walmart dataset before attribute expansion.}
  \label{att}
\end{figure}

\begin{figure}
  \centering
  \includegraphics[width=\linewidth]{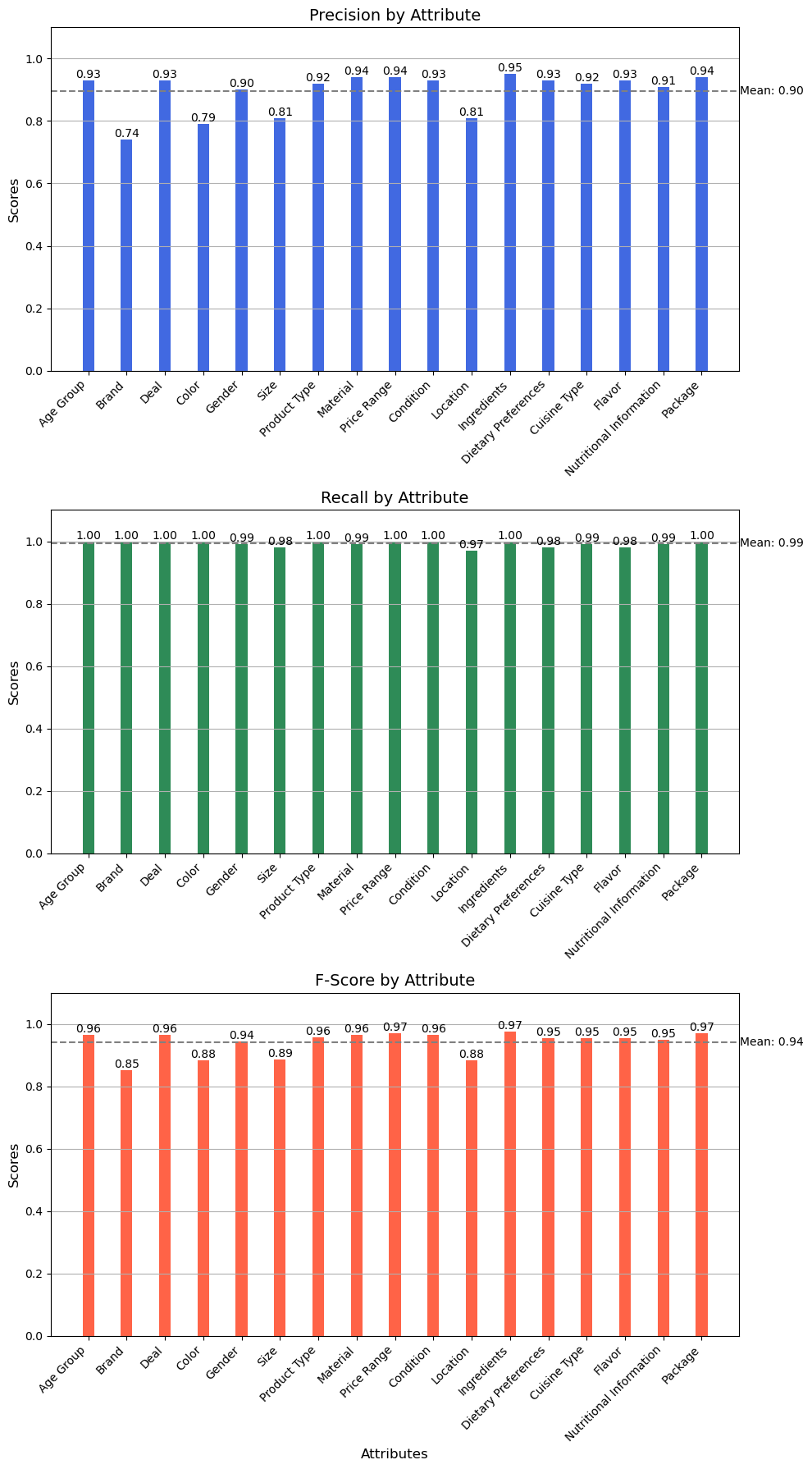}
  \caption{Illustration of the Bert-CRFs NER Method on LLAMA 2.0 annotated data with decorative relationship correction.}
  \label{expan}
\end{figure}

\begin{table*}[ht]
    \caption{The baseline methods' performance across various datasets is presented, with the best results highlighted in bold. All results have achieved statistical significance with 
$p<0.05$.}
    \label{tab:baseline_performance}
    \centering
    \begin{tabular}{c|ccc|ccc|ccc}
        \hline
        Method & \multicolumn{3}{c|}{Walmart} & \multicolumn{3}{c|}{BestBuy} & \multicolumn{3}{c}{CoNLL} \\
        \cline{2-10}
        & Precision & Recall & F-score & Precision & Recall & F-score & Precision & Recall & F-score \\
        \hline
        LSTM-Softmax & 0.8132 & 0.9002 & 0.8544 & 0.8773 & 0.9556 & 0.9149 & 0.9233 & 0.9176 & 0.9204 \\
        GPT2-Softmax & 0.8354 & 0.9346 & 0.8824 & 0.8875 & 0.9608 & 0.9226 & 0.9205 & 0.9054 & 0.9129 \\
        BERT-Softmax & 0.8573 & 0.9654 & 0.9080 & 0.8795 & 0.9654 & 0.9201 & 0.9255 & 0.9277 & 0.9266 \\
        \hline
        LSTM-CRFs & 0.8143 & 0.9449 & 0.8750 & 0.8865 & 0.9520 & 0.9179 & 0.9114 & 0.9253 & 0.9183 \\
        GPT2-CRFs & 0.8496 & 0.9776 & 0.9094 & 0.8954 & 0.9645 & 0.9287 & 0.9024 & 0.8875 & 0.8949 \\
        BERT-CRFs & 0.8674 & 0.9853 & 0.9227 & 0.9033 & 0.9678 & 0.9345 & \textbf{0.9457} & \textbf{0.9255} & \textbf{0.9355} \\
        \hline
        LSTM-CRFs-DRC & 0.8653 & 0.9675 & 0.9132 & 0.9074 & 0.9575 & 0.9319 & 0.9143 & 0.9098 & 0.9120 \\
        GPT2-CRFs-DRC & 0.8829 & 0.9856 & 0.9316 & 0.9144 & 0.9658 & 0.9397 & 0.9065 & 0.8907 & 0.8986 \\
        BERT-CRFs-DRC & \textbf{0.8984} & \textbf{0.9987} & \textbf{0.9457} & \textbf{0.9305} & \textbf{0.9667} & \textbf{0.9484} & 0.9304 & 0.9221 & 0.9262 \\
        \hline
    \end{tabular}
\end{table*}

\subsection{Experimental Results of Attribute Extraction}

We present the comparative performance of various baseline methodologies on the Walmart, BestBuy, and CoNLL datasets in Table \ref{tab:baseline_performance}. The BERT-CRFs-DRC method demonstrated superior performance, achieving the highest precision and recall scores in the Walmart and BestBuy datasets, with F-scores of \(0.9457\) and \(0.9484\) respectively. This reflects its robustness in e-commerce specific entity recognition tasks. However, in the CoNLL dataset, a general NER benchmark, the addition of the DRC module did not enhance the performance of the BERT-CRFs method. This suggests that the DRC's focus on decorative relationships yields benefits specifically in the context of e-commerce search scenarios, underscoring its specialized efficacy in these environments.

Statistical significance tests affirm the robustness of our results with all p-values falling below the \(0.05\) threshold. The distinct performance patterns across datasets underscore the contextual sensitivity of the DRC mechanism: it leverages deep, domain-specific relational insights in e-commerce datasets but does not necessarily translate to generalized NER tasks, where such decorative relationships may be less prevalent or absent.

The analysis of individual attribute performance, as depicted in Figure \ref{att}, indicates that the extraction of 'Brand', 'Color', and 'Size' attributes is particularly challenging for NER systems. This complexity can be attributed to the broad and diverse range of terms associated with these attributes. Nonetheless, the overall F-score achieved is noteworthy, standing at 0.91, with an almost perfect recall rate of 1.00. Such results underscore the efficiency of the proposed model in identifying all pertinent attributes, thereby ensuring that users are provided with a complete array of product choices.

\subsection{Evaluation of LLAMA 2.0 Annotation Methodology}
\label{sec:llama_evaluation}

A thorough human evaluation was carried out to assess the efficacy of the LLAMA 2.0 annotation methodology. Our approach aimed to minimize errors and mitigate subjective judgment by implementing a binary satisfaction metric. Specifically, given a query along with its attributed value, evaluators were asked to provide a binary response—'yes' or 'no'—indicative of their satisfaction with the annotation's accuracy. This binary response system helped us derive the Human Satisfactory Rate (HSR), a clear and objective measure of the annotation's quality and reliability. The HSR thus serves as an integral indicator of LLAMA 2.0’s performance in delivering annotations that align with human perception and understanding.
 The outcomes are documented in Table \ref{tab:human_satisfaction}. The evaluation reveals that 95.53\% of LLAMA 2.0's responses conform to the expected format criteria, indicating that the system reliably returns results with accurate attribute scores derived from the queries. Furthermore, 75.23\% of all queries successfully undergo LLAMA 2.0's review mechanism, demonstrating its effective self-checking capabilities.

These findings highlight that LLAMA 2.0—and by extension, other LLMs utilizing our proposed procedure—yields annotation results that align closely with human performance. This attests to the robustness and reliability of the system in executing annotation tasks.

\begin{table}[htbp]
\centering
\caption{Human Satisfactory Rate (HSR) by Attribute}
\begin{tabular}{l|c}
\hline
\textbf{Attribute} & \textbf{HSR (\%)} \\
\hline
Age Group & 100.0 \\
Brand &  89.0\\
Deal & 100.0 \\
Color & 98.2 \\
Gender & 99.4 \\
Size & 94.6 \\
Product Type & 90.3 \\
Material & 94.3 \\
Price Range & 92.5 \\
Condition & 100.0\\
Location & 98.1 \\
Ingredients & 96.4 \\
Dietary Preferences & 99.8 \\
Cuisine Type & 92.4 \\
Flavor & 96.3 \\
Nutritional Information & 94.3 \\
Package & 92.9 \\
\hline
\end{tabular}
\label{tab:human_satisfaction}
\end{table}

We further demonstrate the efficacy of our proposed method through its performance on the original Walmart dataset. The results, illustrated in Figure \ref{expan}, exhibit a commendable F-score of 0.94. Notably, the method enhances the performance in previously challenging aspects such as brand and color recognition. These empirical results underscore the robustness and industrial applicability of our approach.

\section{Deployment}

Since September 4, 2023, Walmart's Sponsored Product Search has been leveraging our advanced NER algorithm to improve customer search experience. This implementation is aimed at enhancing the precision of search results and advertisement targeting, thereby increasing user satisfaction and engagement.

The integration was executed with precision, ensuring the NER system's capabilities are fully utilized to understand and classify user queries for optimized search outcomes. Early indicators from this deployment suggest a boost in key performance metrics, such as higher click-through rates and reduced instances of search abandonment.

This development has also empowered advertisers to tailor their campaigns more effectively to meet user interests, leading to a more dynamic and responsive shopping experience.

The ongoing application of this AI-driven technology is a clear indicator of our commitment to innovation in e-commerce, with continuous assessments planned to gauge its enduring impact on Walmart's service quality and customer satisfaction.

\section{Conclusion}
In this work, we presented a comprehensive framework that integrates BERT, CRFs, and LLMs to significantly enhance attribute recognition in e-commerce search scenarios. Our innovative decorative relation correction mechanism utilizes the intrinsic relationships between product types and attributes, further sharpening the precision of attribute extraction. We have rigorously tested our approach against multiple datasets, including those from Walmart and BestBuy, as well as the general NER dataset CoNLL, and have observed substantial improvements in performance metrics.

Furthermore, the successful two-month deployment within Walmart's Sponsored Product Search serves as a testament to the model's practical relevance and efficiency in a real-world setting. The utilization of LLMs for data annotation marks a pivotal step towards scaling the model, allowing for the seamless integration of a broader spectrum of attributes while minimizing the need for manual oversight.

In conclusion, our approach not only advances the state-of-the-art in attribute recognition for e-commerce but also opens new avenues for leveraging language models to facilitate continuous improvement and expansion in the domain. The implications of this work extend beyond immediate commercial applications, offering insights and methodologies that can be adapted to a variety of NER tasks across different industries.

\section{Limitations}
While our work has achieved significant improvements in both offline and online metrics, there remain several promising directions for future improvement. Firstly, the recognition results for certain attributes, such as brand, have exhibited a relatively lower level of satisfaction (as shown in Table \ref{tab:human_satisfaction}). One viable approach to address this issue involves enhancing our prompts and conducting additional fine-tuning. Secondly, our present methodology encounters challenges when dealing with queries that encompass multiple product types. This issue could potentially be resolved by implementing grouping techniques to segregate these complex queries \cite{lingraph}.
\bibliographystyle{ACM-Reference-Format}
\bibliography{main}


\end{document}